\documentclass[sigconf]{acmart}
\usepackage{amsmath}
\usepackage{algorithm}
\usepackage{algorithmic}
\usepackage{enumitem}
\usepackage{listings}
\usepackage{setspace}

\definecolor{lightgray}{gray}{0.93}

\AtBeginDocument{%
  }

\copyrightyear{2025}
\acmYear{2025}
\setcopyright{cc}
\setcctype{by-nc-sa}
\acmConference[ICAIF '25]{6th ACM International Conference on AI in Finance}{November 15--18, 2025}{Singapore, Singapore}
\acmBooktitle{6th ACM International Conference on AI in Finance (ICAIF '25), November 15--18, 2025, Singapore, Singapore}\acmDOI{10.1145/3768292.3770433}
\acmISBN{979-8-4007-2220-2/2025/11}

\newcommand{\sharedaffiliation}{%
  \institution{Asian Institute of Digital Finance, National University of Singapore}
  \country{Singapore}}

\begin{document}
\title{FAITH: A Framework for Assessing Intrinsic Tabular Hallucinations in Finance}
\author{Mengao Zhang\textsuperscript{\textdagger}}
\email{mengaoz@nus.edu.sg}
\affiliation{\sharedaffiliation}

\author{Jiayu Fu\textsuperscript{*}}
\email{jiayu.fu@u.nus.edu}
\affiliation{\sharedaffiliation}

\author{Tanya Warrier\textsuperscript{*}}
\email{tanya.warrier@u.nus.edu}
\affiliation{\sharedaffiliation}

\author{Yuwen Wang}
\email{wangyuwen@u.nus.edu}
\affiliation{\sharedaffiliation}

\author{Tianhui Tan}
\email{tant@nus.edu.sg}
\affiliation{\sharedaffiliation}

\author{Ke-Wei Huang}
\email{huangkw@comp.nus.edu.sg}
\affiliation{\sharedaffiliation}

\renewcommand{\shortauthors}{Zhang, et al.}

\begin{abstract}
Hallucination remains a critical challenge for deploying Large Language Models (LLMs) in finance. Accurate extraction and precise calculation from tabular data are essential for reliable financial analysis, since even minor numerical errors can undermine decision-making and regulatory compliance. Financial applications have unique requirements, often relying on context-dependent, numerical, and proprietary tabular data that existing hallucination benchmarks rarely capture. In this study, we develop a rigorous and scalable framework for evaluating intrinsic hallucinations in financial LLMs, conceptualized as a context-aware masked span prediction task over real-world financial documents. Our main contributions are: (1) a novel, automated dataset creation paradigm using a masking strategy; (2) a new hallucination evaluation dataset derived from S\&P 500 annual reports;\textsuperscript{1} and (3) a comprehensive evaluation of intrinsic hallucination patterns in state-of-the-art LLMs on financial tabular data. Our work provides a robust methodology for in-house LLM evaluation and serves as a critical step toward building more trustworthy and reliable financial Generative AI systems.
\end{abstract}


\maketitle
\begingroup
\renewcommand{\thefootnote}{\fnsymbol{footnote}}
\footnotetext{* Work done during internship at the Asian Institute of Digital Finance, National University of Singapore.}
\footnotetext{\textsuperscript{\textdagger} Correspondence.}
\footnotetext{\textsuperscript{1} https://github.com/ZHANG-MENGAO/FAITH }

\endgroup

\section{Introduction}

While Large Language Models (LLMs) are rapidly transforming financial services through capabilities such as automated information extraction \cite{cao2024ecc} and client-facing chatbots \cite{murtaza2025implementing}, their deployment also introduces substantial risks, among which hallucination poses serious threats to decision-making and stakeholder trust  \cite{beyondCutoff, deficiencyInFinance, reduHalluFin, HalluMiniFram, halluMiniJourney}. In response, regulators such as the Monetary Authority of Singapore (MAS) have underscored the need for robust model risk management frameworks tailored to advanced AI applications \cite{MAS2024AI}. Despite growing regulatory scrutiny, systematic methods for evaluating and mitigating hallucinations in financial contexts remain largely undeveloped. As a foundational step, this study proposes a scalable, comprehensive framework for evaluating and analyzing LLM hallucinations in finance, quantifies how increasing information‐extraction complexity exacerbates hallucination errors, and offers actionable guidance for researchers and practitioners. 

Recent empirical studies have begun to reveal that hallucinations in finance exhibit domain-specific characteristics compared with those observed in general-domain NLP benchmarks. For instance, \citet{beyondCutoff} find that LLMs’ factual accuracy on historical data varies across company size and financial periods, while \citet{deficiencyInFinance} and \citet{reduHalluFin} demonstrate that even state-of-the-art models frequently misinterpret financial terms or misreport numerical values. Complementary work such as \citet{halluMiniJourney} outlines stages toward hallucination-minimized systems for financial decision-makers. Together, these studies underscore the urgency of domain-specific evaluation frameworks that capture the unique reasoning and numerical dependencies underlying financial data. However, a systematic, fine-grained benchmark that directly measures hallucinations in finance remains lacking.

This paper focuses on intrinsic hallucinations, a form of LLM error where the generated output contradicts or misrepresents the input financial context \cite{halluLens}. Many financial tasks require precise extraction, summarization, calculation, and reasoning based on structured statements such as balance sheets or income tables. Errors in these processes can significantly undermine analytical integrity. While intrinsic hallucinations are already concerning in such settings, reliance on external knowledge or web retrieval amplifies risks. Undetected hallucinations can propagate through automated reporting pipelines or investment algorithms, potentially resulting in misleading insights, compliance violations, or financial losses. From a practical standpoint, even modest hallucination rates—such as 10–20\% in complex financial reasoning—can have outsized impacts when scaled across high-stakes decisions in portfolio management or regulatory reporting.

Yet, evaluating hallucinations in a systematic and robust manner remains a non-trivial challenge due to the scarcity of finance-specific benchmarks. Existing datasets are predominantly derived from general-domain texts such as Wikipedia \cite{anah2, halluLens}, whereas financial applications involve numerically grounded, context-dependent data and specialized terminology. More importantly, there is a clear need for an automated and scalable approach to creating evaluation datasets for hallucination detection in real-world financial settings. Manual annotation is resource-intensive and can hardly keep up with evolving LLM behaviors or the diversity of proprietary financial data. 

To address these gaps, we develop FAITH (A Framework for Assessing Intrinsic Tabular Hallucinations in Finance) - an automated and scalable approach for constructing intrinsic hallucination evaluation tasks directly from real financial annual reports, where ground-truth numerical values are explicitly available. We address a core challenge in financial LLM applications: accurately identifying, calculating, and presenting numerical values based on financial statements (input context). We introduce a novel context-aware masked span prediction task, where numerical values from real annual reports are masked and LLMs are prompted to recover them. To ensure reliable evaluation, we develop a novel filtering method that selects only those values with a unique, consistent, and answerable ground truth. A central contribution is our taxonomy of four mutually exclusive and exhaustive financial reasoning types: \textit{Direct Lookup}, \textit{Comparative Calculation} (The value is a function of the same variable in different periods, e.g., year-over-year growth), \textit{Bivariate Calculation} (the value is a function of two variables, e.g., gross margin), and \textit{Multivariate Calculation} (more complex dependencies). This classification enables structured evaluation across reasoning complexity. Using our approach, we build a large-scale benchmark dataset from S\&P 500 annual reports and evaluate leading LLMs on their intrinsic hallucination rates.

\textbf{Empirically, our results reveal a stratified hierarchy of model reliability in financial reasoning.} Proprietary frontier models such as Claude-Sonnet-4 and Gemini-2.5-Pro achieve high overall accuracy, yet even these systems exhibit 10–20\% error rates on multi-step numerical reasoning. In contrast, smaller open-source models often fail catastrophically on complex calculations, scoring near zero in multivariate scenarios. This degradation with reasoning complexity highlights an urgent limitation of current architectures—the difficulty of maintaining factual consistency across chained numerical operations. Importantly, such residual hallucination rates, though seemingly modest, could translate into substantial financial misjudgments in investment analytics, credit risk assessment, or regulatory compliance when scaled to production systems.

\begin{figure*}[t]
    \centering
    \includegraphics[width=0.85\textwidth]{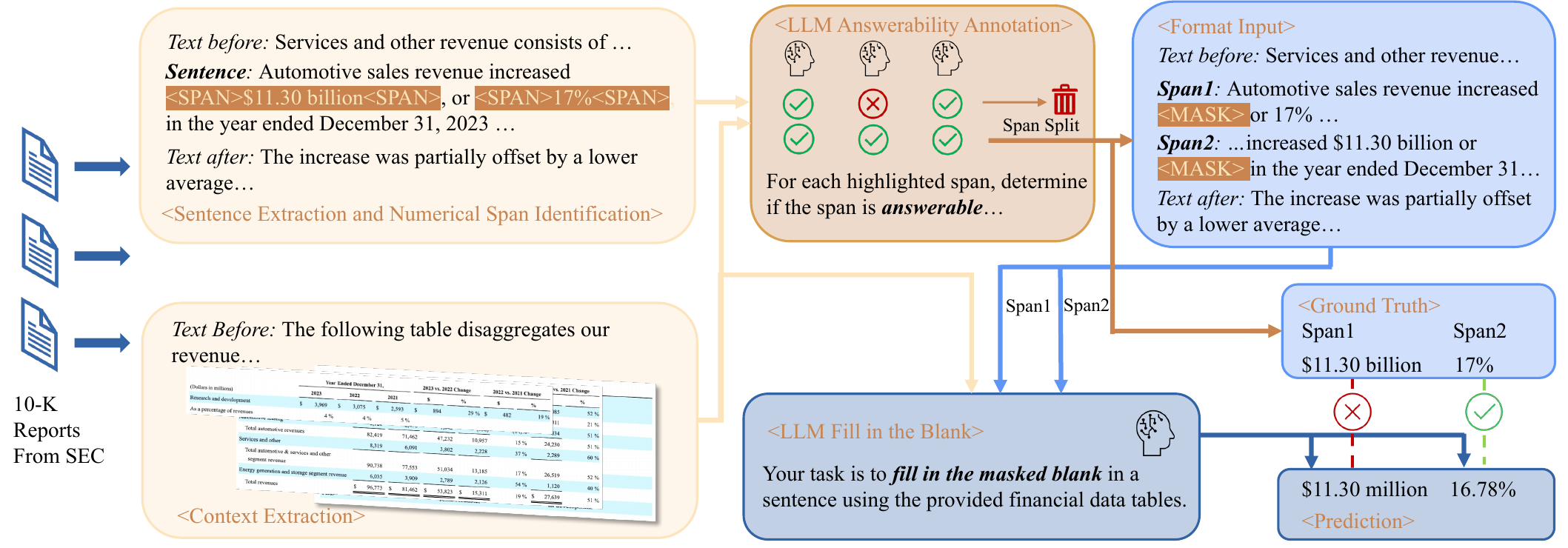}
    \caption{Illustration of the task definition and data processing.}
    \label{fig:task_definition}
\end{figure*}

Our main contributions are as follows: 
\begin{itemize}
    \item We propose a novel and scalable paradigm for dataset creation based on a masking strategy, enabling automatic construction of evaluation datasets for both public and proprietary financial documents.
    \item We release a new dataset for evaluating intrinsic hallucinations in finance, with varying reasoning complexities. 
    \item  We conduct a comprehensive analysis of intrinsic hallucination patterns in state-of-the-art LLMs on financial tabular data, including a detailed breakdown by reasoning type.
\end{itemize}

\section{Related Works}
\subsection{Hallucination Evaluation Datasets}

LLM hallucinations can be categorized as extrinsic or intrinsic, based on the reference source \cite{halluLens}. The evaluation of extrinsic hallucinations assesses whether a model's output, generated from its internal knowledge alone, aligns with established world facts. Benchmarks in this category \cite{Min2023FActScore, li-etal-2023-halueval, lin-etal-2022-truthfulqa} generally evaluate models against static, open-domain corpora like Wikipedia. Financial LLMs, however, operate in a domain where information is highly time-sensitive, context-intensive, and often proprietary, requiring evaluation methods that prioritize contextual accuracy over pre-trained, parametric knowledge \cite{beyondCutoff, deficiencyInFinance, reduHalluFin, HalluMiniFram, halluMiniJourney}.

Therefore, intrinsic hallucination evaluation is more relevant and critical for financial applications, as it directly measures a model's faithfulness to a specific, provided context. While intrinsic evaluation datasets have emerged, especially for Retrieval-Augmented Generation (RAG) applications \cite{RAGTruth, FaithEval, anah2}, those based on general-purpose text, mostly are limited to simple look-up tasks rather than complex reasoning \cite{RAGTruth, azaria2023internalstatellmknows}.

Our work addresses this critical gap, focusing on intrinsic hallucinations within complex financial tabular data—a unique and high-stakes domain largely overlooked by current evaluation methodologies.

\subsection{Tabular Reasoning and Evaluation}

Our focus on intrinsic hallucinations with tables connects to the broader research area of tabular reasoning. Foundational benchmarks such as \texttt{RealHiTBench} \cite{Wu2025RealHiTBench} and \texttt{TableBench} \cite{Wu2025TableBench} have been pivotal in assessing core LLM skills like numerical reasoning. In the more intricate financial domain, specialized benchmarks have emerged to address higher complexity. Datasets like \texttt{FinQA} \cite{Chen2021FINQA} and \texttt{TAT-QA} \cite{Zhu2021TATQA} test joint reasoning over tables and text, while others focus on navigating complex hierarchical structures \cite{zhao2022multihiertt,Krumdick2024Bizbench,Wang2025FinTagging}.

While these benchmarks have significantly advanced LLM reasoning capabilities, their evaluation scope is primarily focused on the correctness of the final answer. This focus on accuracy creates a critical blind spot: as task complexity increases, so does the risk that a model generates a correct-seeming output through unfaithful reasoning or hallucination. Addressing this gap requires moving beyond traditional question-answering formats to develop scalable methodologies that can directly probe for such intrinsic errors. Our study proposes a novel framework to rigorously and automatically evaluate intrinsic hallucination in financial tabular reasoning.

\subsection{Automatic Dataset Construction}
The high cost and limited scalability of manual annotation have driven the development of automated evaluation dataset construction methods. Some approaches achieve fine-grained analysis by decomposing generated content into atomic facts for verification \cite{Min2023FActScore, Akbar2024HalluMeasure, Hu2024RefChecker}, while others leverage the inherent structure of relational data to automatically generate complex, verifiably question-answer pairs \cite{Oh2024ERBench}. Adversarial test cases can be dynamically created through data perturbation \cite{Yu2024ReEval}. Although using LLMs as annotators may introduce biases \cite{Zheng2023Judging}, recent work shows that carefully designed automated pipelines can yield high-quality annotations, sometimes rivaling human performance \cite{Luo2024HalluDial}. We build on these advancements by adapting and extending automated benchmarking paradigms to the unique challenges of financial tabular analysis.

\section{Task Definition}
In this section, we focus on the core requirement of retrieving information from context and providing accurate answers in complex financial scenarios, which is a common need for financial LLM applications. We formulate the task as a context-aware masked span prediction over tabular financial data, for the goal of intrinsic hallucination evaluation. Figure \ref{fig:task_definition} demonstrates key flow.

\subsection{Problem Formulation}
Let a financial document $D$ be composed of a set of structured tables $\mathcal{T}$ and a body of text. This text is partitioned into two distinct, non-overlapping sets:
\begin{itemize}[leftmargin=1.5em]
    \item A set of explanatory pre-texts $\mathcal{P}$, where each pre-text $P_T \in \mathcal{P}$ introduces the table $T\in \mathcal{T}$.
    \item A sequence of general sentences $S = (s_1, s_2, ..., s_N)$, which constitutes the remaining text of the document. 
\end{itemize}
Within a given sentence $s_i \in S$, we identify a set of non-overlapping spans of interest, $\mathcal{M}_i=\{m_{i,1}, m_{i,2}, ..., m_{i,k}\}$, based on specific criteria. A task instance is constructed by selecting a single span $m_{i,j} \in \mathcal{M}_i$ and replacing its content with a \texttt{[MASK]} token, producing a corrupted sentence $\tilde{s_i}$. 

The objective of the model $f$ is to recover the original content of the masked span $m_{i,j}$. The model is conditioned on the corrupted sentence and a context set $C_i$:
\begin{equation}
    \hat{m}_{i,j} = f(\tilde{s_i}, C_i)
\end{equation}

The context $C_i=\{\mathcal{T}, \mathcal{P}, s_{i-1}, s_{i+1}\}$ includes the tables, their pre-text, and the immediately preceding and succeeding sentences. The goal is for the predicted span $\hat{m}_{i,j}$ to be matched with the original span $m_{i,j}$

\subsection{Masking Criteria}
To ensure the task setup yields meaningful and unbiased hallucination evaluation, three key assumptions must hold for each masked span. These assumptions guide our masking strategy and directly impact the reliability of evaluation outcomes:

\begin{itemize}[leftmargin=1.5em]
    \item \textbf{Uniqueness:} The masked span must have a unique correct answer, preventing multiple plausible completions.
    \begin{itemize}[leftmargin=1em]
        \item \textit{Example:} A sentence like "The company’s [MASK] has improved" would be excluded, as several financial metrics such as "revenue," "profit margin," or "cash flow" could all be reasonable answers.
    \end{itemize}

     \item \textbf{Consistency:} The ground-truth span must be consistent with the context, avoiding internal misalignments within the source document.
    \begin{itemize}[leftmargin=1em]
        \item \textit{Example:} If the table reports operating income as \textit{\$500 million}, the masked sentence should not mistakenly state "Operating income was \textit{\$500 thousand}" as the to-be-masked truth. Evaluating based on the incorrect masked content would unfairly penalize a model that correctly aligns with the table.
    \end{itemize}
    
    \item \textbf{Answerability:} The masked span must be inferable from the provided context, ensuring the task is solvable by an LLM.
    \begin{itemize}[leftmargin=1em]
        \item \textit{Example:} If the sentence states "The company's revenue increased by [MASK] in 2024 compared to 2023," both revenue figures for 2024 and 2023 should be present in the tables or described in the surrounding text; otherwise, the span is not answerable and should be excluded from evaluation.
    \end{itemize}
\end{itemize}

To enforce these assumptions and ensure reliable evaluation, we adopt the following design:

\begin{enumerate}[leftmargin=1.5em]
    \item \textbf{Numeric Span Selection:} We restrict masking to numeric spans that include units or verbal scales (e.g., "million," "USD"), ensuring that the masked content is both specific and uniquely recoverable. We also implement normalization strategies for evaluation to account for equivalent numeric representations.
    
    \item \textbf{Reliable Source Documents:} We use company annual reports (i.e., 10-K reports) as the data source, minimizing the likelihood of contradictions due to their regulatory rigor and editorial consistency.
    
    \item \textbf{Answerability Annotation:} We utilize LLMs to annotate the answerability of the spans and conducted a comprehensive pilot study by comparing human annotations with LLM-generated annotations, demonstrating that LLMs can reliably support the answerability labeling process.
\end{enumerate}

\subsection{Robust Evaluation}
\label{sec:robust_eval}

\begin{algorithm}[t]
\caption{Precision-Relaxed Matching with Unit Groups}
\label{alg:matching}
\begin{algorithmic}[1]
\REQUIRE Ground truth span $m$, predicted span $\hat{m}$, unit groups $\mathcal{U}$
\ENSURE $is\_numeric\_match$ and $is\_unit\_match$

\STATE Let $\mathcal{U}_{\text{scales}} \subset \mathcal{U}$ be the subset of scale units
\STATE $(v_m, p_m) \gets \text{NormalizeNumber}(m, \mathcal{U}_{\text{scales}})$
\STATE $(v_{\hat{m}}, p_{\hat{m}}) \gets \text{NormalizeNumber}(\hat{m}, \mathcal{U}_{\text{scales}})$
\item[]\hspace{2.2cm}\small\textit{// Normalize to base value and determine precision}
\item[]

\STATE $p_{coarsest} \gets \max(p_m, p_{\hat{m}})$
\STATE $is\_numeric\_match \gets (\lfloor v_m / p_{\min} \rceil = \lfloor v_{\hat{m}} / p_{\min} \rceil)$
\item[]\hspace{3.8cm}\small \textit{// Compare numbers after rounding}
\item[]

\STATE $U_m \gets \text{ExtractUnits}(m, \mathcal{U} \setminus \mathcal{U}_{\text{scales}})$
\STATE $U_{\hat{m}} \gets \text{ExtractUnits}(\hat{m}, \mathcal{U} \setminus \mathcal{U}_{\text{scales}})$
\item[]\hspace{2.5cm}\small \textit{    // Extract non-scale unit groups from each span}
\item[]

\STATE $is\_unit\_match \gets (U_m \subseteq U_{\hat{m}})$
\item[]\hspace{3.5cm}\small \textit{    // Check for set equality of unit groups}
\item[]
\RETURN $is\_numeric\_match, is\_unit\_match$
\end{algorithmic}
\end{algorithm}








Our evaluation protocol is designed to robustly handle the nuances of numerical text, ensuring that valid predictions are not penalized due to formatting or semantic variations. We specifically account for two potential sources of ambiguity where a simple string comparison would prove insufficient:

\begin{enumerate}[leftmargin=1.5em]
    \item \textbf{Compromised Uniqueness:} In certain contexts, a masked span could be correctly expressed in multiple values, violating the Uniqueness assumption. For instance, in \textit{"the company's revenue increased by [MASK],"} both a percentage change (\textit{"10\%"}) and an absolute value (\textit{"\$5 million"}) could be factually correct based on the source table.
    \item \textbf{Formatting Variations:} A single numeric value can be written in many equivalent string formats (e.g., \textit{\$1,230 million} vs. \textit{USD 1.23 billion}). A simple string match would incorrectly penalize valid predictions.
\end{enumerate}

To address the first challenge, we guide the model by incorporating a \textbf{hint} into the prompt that specifies the expected value type (e.g., percentage or absolute value), thereby restoring uniqueness. To solve the second, we introduce a \textbf{precision-relaxed evaluation protocol} that normalizes and compares the numeric predictions. This protocol is detailed in Algorithm~\ref{alg:matching} and consists of two main components:

\paragraph{Numeric Matching.} This process first normalizes the ground-truth span $m$ and the predicted span $\hat{m}$ into base values. It parses numbers and scale indicators (e.g., ``million,'' ``billion'') to derive a base number $v$ (e.g., ``1.23 billion'' becomes $1.23 \times 10^9$). It then determines the numerical \textbf{precision} from the least significant digit. For instance, the number 1,230,000 has a precision $p$ of $10^4$, as its last non-zero digit is in the ten thousands place. A numeric match is declared if the ground-truth and predicted values are equal after being rounded to the coarser of their two precisions.

\paragraph{Unit Matching.} We define a set of \textbf{unit groups} $\mathcal{U}$, where each group contains aliases for a single unit (e.g., \{\$, USD, dollars\}, \{mil, million, M\}). We extract the set of unit groups present in each span by greedily matching the longest aliases first. For non-scale units, matching is successful only if all groups found in $m$ are present in $\hat{m}$
\subsection{Financial Reasoning Scenarios}
\label{sec:scenarios}
To facilitate a fine-grained analysis of model performance, we categorize each masked span based on the complexity of the financial reasoning required to restore its content. We define four distinct scenarios:

\begin{itemize}[leftmargin=1.5em]
    \item[A.]\textbf{Direct Lookup:} The answer can be found by directly extracting a single cell in a table.
    \item[B.]\textbf{Comparative Calculation:} The answer requires a simple calculation on a single metric across different time periods or categories (e.g., computing a year-over-year change).
    \item[C.]\textbf{Bivariate Calculation:} The answer involves a simple calculation between two distinct metrics explicitly present in the table (e.g., computing a financial ratio).
    \item[D.]\textbf{Multivariate Calculation:} The answer requires multi-step reasoning over three or more metrics or a sequence of arithmetic operations.
\end{itemize}

Recognizing that a masked span can sometimes be derived through multiple valid reasoning paths, we do not let LLM determine a fixed scenario label during answerability annotation step. Instead, during evaluation time, we instruct each model to classify its own reasoning process into one of the four scenarios. To establish a reliable scenario classification for our analysis, we aggregate the scenario classifications from all models that correctly predict the span's content. The final scenario for that span is the class with highest frequency, ensuring our analysis is grounded in successful and verifiable reasoning steps from LLM.

\subsection{Pilot Study}

\begin{table}[t]
    \centering
    \caption{Agreement among LLMs and human annotation for Answerability Annotation. "Yes" and "No" represent "answerable" and "unanswerable" respectively. The LLMs used are GPT-4.1, Claude-sonnet-4, and Gemini-2.5-pro.}
    \label{tab:llm_agreement_pilot}
    \begin{tabular}{lrrr}
        \toprule
        & \multicolumn{2}{c}{\textbf{Human}} & \\
        \cmidrule(lr){2-3}
        \textbf{LLM Agreement Pattern} & \textbf{Yes} & \textbf{No} & \textbf{Total} \\
        \midrule
        \textbf{All 3 Agree (Unanimous)} \\
        \quad 3 LLMs - No & 1 & 626 & 627 \\
        \quad 3 LLMs - Yes & 276 & 11 & 287 \\
        \midrule
        \textbf{2-1 Split (Disagree)} \\
        \quad 2 LLMs - Yes, 1 LLM - No & 12 & 23 & 35 \\
        \quad 1 LLM - Yes, 2 LLMs - No & 11 & 164 & 175 \\
        \midrule
        \textbf{Total} & 300 & 824 & 1124 \\
        \bottomrule
    \end{tabular}
\end{table}


To validate our evaluation dataset construction methodology, we conduct a pilot study to assess the feasibility of using LLMs for annotation.

\textbf{Manual Annotation.} We begin by creating a ground-truth dataset. We sample 1,124 text spans from the \textit{"Item 7. Management's Discussion and Analysis"} section of 10-K reports from nine companies across diverse industries (e.g., healthcare, finance, technology). Each span is labeled for answerability by at least two financial experts, with a senior reviewer adjudicating disagreements. This process yields a high inter-annotator agreement (Fleiss' Kappa = 0.905) and results in 300 answerable and 824 unanswerable spans. The annotators also classify the 300 answerable spans into our four financial reasoning scenarios (A: Direct Lookup, B: Comparative, C: Bivariate, D: Multivariate), resulting in a distribution of 212, 58, 22, and 8 respectively, which reflects the natural rarity of more complex reasoning tasks.

\textbf{LLM Annotation.} We then prompt three leading LLMs (GPT-4.1, Claude-Sonnet-4, Gemini-2.5-Pro) with the answerability annotation task. The results, summarized in Table~\ref{tab:llm_agreement_pilot}, show that \textbf{unanimous LLM consensus is an exceptionally strong indicator of answerability correctness}. When all three models agreed, they correctly identified 626 of 627 unanswerable spans (99.8\% accuracy) and 276 of 287 answerable spans (96.2\% accuracy).

This pilot study validates that leveraging unanimous LLM consensus is a highly reliable and scalable strategy for annotating answerability. It is worth noting that we only mask existing text rather than generating new content; thus, using LLMs as annotators does not introduce fabrications. This ensures the soundness of our approach for creating intrinsic hallucination evaluation benchmarks.

\section{Evaluation Dataset}

\subsection{Data Collection and Processing}
Our dataset is built upon publicly available 10-K reports from S\&P 500 companies filed in 2024, ensuring the data reflects real-world financial reporting practices and covers a wide spectrum of industries. We source these documents in XBRL format from the SEC's EDGAR database\footnote{https://www.sec.gov}. The collected filings then undergo several processing steps to extract high-quality numerical claims and their surrounding context.

\textbf{Item 7 Extraction.} From each 10-K filing, we extract "Item 7: Management's Discussion and Analysis of Financial Condition and Results of Operations" (MD\&A). This section is selected for its rich narrative analysis and manageable length, providing a dense source of contextualized financial data. We apply a keyword-based search to locate the MD\&A section, supplementing this automated process with manual curation to ensure data quality.

\begin{table}[t]
    \centering
    \caption{Comparative statistics of the Pilot (human-annotated) and Main (LLM-annotated) dataset splits.}
    \label{tab:dataset_statistics}
    \begin{tabular}{lrr}
        \toprule
        \textbf{Metric} & \textbf{Pilot Split} & \textbf{Main Split} \\
        \midrule
        Number of companies & 9 & 453 \\
        Avg. context length (chars) & 14,148 & 12,843 \\
        Avg. number of tables & 14.9 & 19.2 \\
        Number of sentences & 164 & 1,122 \\
        Number of answerable spans & 300 & 2,406 \\
        \bottomrule
    \end{tabular}
    \normalsize
\end{table}

\textbf{Context and Sentence Extraction.} To avoid overlaps between context and sentence, we first extract tables and their immediate preceding sentence as the explanatory pre-texts. Then we parse the rest of the content into plain text. After which a sentence split is performed using spaCy to get a list of candidate sentences which incorporated custom rules to merge fragments that typically arose as artifacts from the document conversion process. 

\textbf{Numerical Span Identification.} From this sentence corpus, we isolate claims containing numerical data. This is a two-stage process:
\begin{itemize}[leftmargin=1.5em]
    \item \textbf{Initial Detection:} We employ spaCy's Named Entity Recognition (NER) to detect entities such as MONEY, PERCENT, CARDINAL, and QUANTITY.
    \item \textbf{Span Expansion:} To ensure the extracted spans are semantically complete, we expand the initial NER outputs with rules to include relevant currency symbols (e.g., \$) and a comprehensive vocabulary of financial units and scales (e.g., "million," "billion," "per share").
\end{itemize}

\textbf{LLM-based Answerability Annotation.} To create a representable sample, we randomly select 10 sentences from each 10-K report. Following the same setting in pilot study, three popular LLMs are employed for annotating whether each span is answerable given the provided context. A span is retained in the final dataset only if all three models unanimously classified it as answerable.

\subsection{Dataset Statistics}

\begin{table*}
\centering
\small 
\caption{Model Accuracy (\%) on Pilot and Main Splits, including scenario performance. Scenario abbreviations: (A) Direct Lookup, (B) Comparative Calculation, (C) Bivariate Calculation, and (D) Multivariate Calculation.}
\label{tab:full_merged_performance}
\begin{tabular}{l|ccccccc|cccccccc}
\toprule
\textbf{Model} & \multicolumn{7}{c}{\textbf{Pilot Split}} & \multicolumn{7}{c}{\textbf{Main Split}} \\
\cmidrule(lr){2-8} \cmidrule(lr){9-15}
& \textbf{Overall} & \textbf{Value} & \textbf{Unit} & \textbf{A} & \textbf{B} & \textbf{C} & \textbf{D} & \textbf{Overall} & \textbf{Value} & \textbf{Unit} & \textbf{A} & \textbf{B} & \textbf{C} & \textbf{D} \\
& & & & \footnotesize{(n=212)} & \footnotesize{(n=58)} & \footnotesize{(n=22)} & \footnotesize{(n=8)} & & & & \footnotesize{(n=1606)} & \footnotesize{(n=635)} & \footnotesize{(n=135)} & \footnotesize{(n=10)} \\
\midrule
Gemini-2.5-pro & \textbf{95.0} & \textbf{95.7} & \textbf{97.7} & \textbf{96.7} & 91.4 & \textbf{95.5} & \textbf{75.0} & 91.9 & \textbf{97.8} & 93.1 & 91.8 & \textbf{94.0} & \textbf{96.3} & \textbf{90.0} \\
Gemini-2.5-flash & 91.0 & 92.3 & 96.7 & 94.8 & 82.8 & 90.9 & 50.0 & 88.7 & 91.1 & 93.7 & 90.2 & 88.0 & 87.4 & 70.0 \\
Gemini-2.5-flash-lite & 55.7 & 59.0 & 83.0 & 67.0 & 27.6 & 27.3 & 37.5 & 50.2 & 57.9 & 57.8 & 45.8 & 59.4 & 70.4 & 20.0 \\
Claude-sonnet-4 & 93.0 & 93.7 & 97.3 & 96.2 & \textbf{93.1} & 81.8 & 37.5 & \textbf{95.6} & 95.8 & \textbf{98.7} & \textbf{97.0} & 82.6 & 94.1 & 80.0 \\
Claude-haiku-3 & 64.7 & 66.0 & 92.7 & 71.7 & 50.0 & 59.1 & 0.0 & 81.3 & 81.8 & 93.9 & 84.1 & 80.6 & 66.7 & 30.0 \\
GPT-4.1 & 89.7 & 90.3 & 95.3 & 93.9 & 87.9 & 72.7 & 37.5 & 89.2 & 89.5 & 92.6 & 91.1 & 90.4 & 77.8 & 30.0 \\
GPT-4.1-mini & 78.0 & 79.3 & 91.7 & 85.4 & 70.6 & 40.9 & 37.5 & 88.2 & 89.4 & 94.2 & 89.7 & 89.0 & 80.7 & 70.0 \\
GPT-4.1-nano & 31.3 & 43.3 & 62.3 & 36.8 & 19.0 & 22.7 & 0.0 & 70.0 & 71.9 & 92.1 & 71.8 & 72.3 & 53.3 & 10.0 \\
Qwen-3-8B & 20.3 & 32.0 & 44.3 & 24.5 & 10.3 & 13.6 & 0.0 & 30.6 & 35.6 & 36.1 & 27.1 & 35.7 & 54.1 & 0.0 \\
Qwen-3-32B & 68.0 & 70.0 & 96.0 & 80.7 & 34.5 & 45.5 & 37.5 & 73.9 & 76.7 & 83.7 & 73.5 & 76.1 & 81.5 & 40.0 \\
Ministral-8B & 22.3 & 23.0 & 75.7 & 29.7 & 1.7 & 9.1 & 12.5 & 40.8 & 41.8 & 74.7 & 45.3 & 33.2 & 31.9 & 0.0 \\
Mistral-small-24B & 63.7 & 63.7 & \textbf{97.7} & 77.8 & 29.3 & 31.8 & 25.0 & 45.6 & 64.2 & 53.4 & 45.3 & 46.0 & 57.0 & 0.0 \\
Llama-3.1-8B & 27.0 & 29.0 & 68.3 & 35.3 & 6.9 & 9.1 & 0.0 & 47.5 & 47.8 & 85.5 & 47.8 & 49.9 & 43.7 & 10.0 \\
Llama-3.3-70B & 56.0 & 57.0 & 95.3 & 66.0 & 36.2 & 31.8 & 0.0 & 37.0 & 40.9 & 42.5 & 34.7 & 40.9 & 53.3 & 10.0 \\
Gemma-3-12B & 32.0 & 34.3 & 83.7 & 42.5 & 6.9 & 9.1 & 0.0 & 15.2 & 19.7 & 37.7 & 16.3 & 12.8 & 17.0 & 0.0 \\
Gemma-3-27B & 37.3 & 39.0 & 91.0 & 45.7 & 20.7 & 13.6 & 0.0 & 33.8 & 37.3 & 51.0 & 31.7 & 38.6 & 43.7 & 0.0 \\
\bottomrule
\end{tabular}
\end{table*}
Table \ref{tab:dataset_statistics} presents a statistical overview of our dataset, which is composed of a human-annotated Pilot Split and a larger, LLM-annotated Main Split covering 453 S\&P 500 companies after processing and filtering.

A key characteristic of our dataset is its realistic complexity. With an average context length exceeding 12,800 characters and an average of 19.2 tables per document, the dataset is designed to mirror the information-dense nature of real-world financial reports. This substantial context poses a significant challenge for LLMs, providing a robust testbed for evaluating their reasoning and retrieval capabilities under practical, real-world conditions.

\section{Experiments}

\subsection{Experimental Setup}
We evaluate a range of state-of-the-art open-source and proprietary LLMs on our dataset's pilot and main splits. During evaluation, we prompt the models to generate a step-by-step rationale before predicting the final masked value and to self-classify their reasoning process into one of the four financial scenarios (A-D). The detailed prompt is available in Appendix \ref{sec:ans_prompt}.

Accuracy is used as the primary evaluation metric, computed for overall predictions and separately for the numeric value and unit components. In addition, we report the accuracy stratified by financial scenario. For the pilot split, scenario labels are assigned by human annotators. For the main split, we use the strategy stated in section \ref{sec:scenarios}. This stratification allows for a detailed analysis of error types.

All models are run with a temperature of 0 to ensure reproducibility. Proprietary models are accessed via their official APIs, while open-source models are run on 4 H200 GPUs using the LLaMA-Factory framework \cite{llamafactory}. We implement a retry mechanism (up to 3 attempts) to handle instances where a model failed to produce a valid JSON output.

\subsection{Experiment Results}

This section presents a detailed analysis of the performance of various LLMs on our financial hallucination benchmark. The comprehensive results, detailed in Table~\ref{tab:full_merged_performance}, facilitate an objective assessment of current model capabilities and limitations in finance. Our findings reveal a clear stratification of model performance, underscore the direct relationship between reasoning complexity and hallucination risk, and identify dominant error patterns that highlight specific model weaknesses.

\subsubsection{A Quantifiable Hierarchy of Model Reliability}

The results demonstrate a distinct performance hierarchy among the evaluated models.

A top tier of proprietary models, led by \textbf{Claude-Sonnet-4} (95.6\% on Main Split) and \textbf{Gemini-2.5-Pro} (91.9\% on Main Split), exhibits high overall accuracy. Their strong performance, which remains largely consistent between the Pilot and Main splits, validates the robustness of our benchmark. However, the 4-8\% error rate, while low relative to other models, remains a significant consideration for financial applications where precision is non-negotiable.

Below this top tier, we observe a second group of capable models, including \textbf{GPT-4.1} (89.2\%) and \textbf{GPT-4.1-mini} (88.2\%), which deliver respectable but demonstrably lower accuracy. A further sharp decline is evident in the majority of other models, particularly smaller open-source variants. Models such as \textbf{Llama-3.1-8B} (47.5\%) and \textbf{Qwen-3-8B} (30.6\%) exhibit high error frequencies, rendering them unsuitable for tasks requiring financial fidelity.

\subsubsection{Reasoning Complexity as the Primary Performance Differentiator}
The analysis of performance across the four reasoning scenarios (A-D) reveals that task complexity is the most significant factor influencing model reliability.

\paragraph{Degradation in Multi-step Scenarios.} While most models perform adequately on \textbf{Direct Lookup (A)}, accuracy systematically decreases as tasks require calculation and logical inference. The most pronounced performance drop occurs in the \textbf{Multivariate Calculation (D)} scenario, which serves as a stress test for multi-step reasoning. On this task, a significant number of models, including several with relatively higer parameter counts (e.g., Llama-3.3-70B, Mistral-small-24B), score at or near 0.0\%. This indicates a fundamental breakdown in their ability to reason under complex context, leading to the fabrication of outputs.

\paragraph{Resilience in Top-Tier Models.} In contrast, the top-performing models demonstrate notable, albeit imperfect, resilience on complex tasks. \textbf{Gemini-2.5-Pro} is particularly robust in the \textbf{Multivariate Calculation (D)} scenario, achieving 90.0\% accuracy on the Main Split. \textbf{Claude-Sonnet-4} also performs strongly at 80.0\%. While these results are promising and highlight advanced reasoning capabilities, a 10-20\% failure rate on the most complex calculations represents a critical barrier. This difficulty in maintaining factual consistency through multi-step logic remains a primary vector for intrinsic hallucinations.
\begin{table}[t]
    \centering
    \caption{Key financial attributes for Mohawk Commons used in the masked value inference task. \textit{(Table headers and rows have been truncated for clarity.)}}
    \small
    \begin{tabular}{@{}l r r r l@{}}
        \toprule
        \textbf{Investment} & \textbf{Ownership \%} & \textbf{Debt (\$M)} & \textbf{Rate} & \textbf{Maturity} \\
        \midrule
        Gotham & 49\% & \$8.5 & 8.36\% & Mar 2024 \\
        Mohawk Commons & 18.1\% & \$7.2 & 5.80\% & Mar 2028 \\
        \midrule
        \textbf{Total} & & \textbf{\$188.8} & & \\
        \bottomrule
    \end{tabular}
    \label{tab:ownership-debt}
\end{table}
\subsubsection{Case Study}
Beyond aggregate metrics, we perform a qualitative analysis to diagnose the models' specific failure modes. Our analysis reveals several recurring patterns, with one of the most significant being scale error. This occurs when a model correctly identifies a numerical value but fails to associate it with the proper magnitude (e.g., reporting "\$150" instead of "\$150 million"). Correcting for this single error type in Llama-3.3-70B's outputs, for instance, would boost its value accuracy from 37.0\% to 57.7\%, highlighting this as a critical vulnerability in contextual numerical grounding.

However, more profound failures stem from a deficient understanding of financial concepts that require multi-step reasoning across different data sources. To illustrate these deeper challenges, we now present a representative detailed case study on a sample that required latent variable inference from both tabular and textual context. The target span was the equity investment value (\$20.2 million) in the following scenario:
\begin{quote}
    \textit{On January 27, 2023, Fund V acquired a 90\% interest in an unconsolidated venture for \textbf{\$20.2 million}, which purchased a shopping center, Mohawk Commons, located in Schenectady, New York, for \$62.1 million, inclusive of transaction costs.}
\end{quote}

Table~\ref{tab:ownership-debt} contains the relevant details for Mohawk Commons, including the 18.1\% ownership and the pro-rata share of mortgage debt (\$7.2M).

\textbf{Model Reasoning Patterns:}
Only Gemini-2.5-Pro accurately inferred the masked value. Its reasoning chain involved several inferential steps: The model first identified the relevant table entry (Mohawk Commons) that matched the acquisition described in the text. The model then inferred the total mortgage debt on the asset by dividing the pro-rata debt (\$7.2\,\text{M}) by the ownership percentage (18.1\%), yielding \$39.7\,\text{M}. Subtracting this value from the purchase price (\$62.1\,\text{M}) gave the total equity (\$22.4\,\text{M}), to which the model applied the Fund’s 90\% interest, ultimately arriving at the correct equity investment value of \$20.2\,\text{M}.

In contrast, GPT-4.1 and Claude-Sonnet-4 failed to utilize the debt information from the table. Instead, both models based their prediction solely on the information in the sentence, by simply calculating 90\% of the purchase price (\(62.1\,\text{M} \times 0.90 = 55.9\,\text{M}\)). This neglects the mortgage debt and indicates insufficient integration of tabular data.

\textbf{Implications for Multimodal Reasoning: }
This example reveals a key limitation in current LLM capabilities: the inference of latent variables that require synthesizing information across modalities and reasoning over implicit relationships. Only Gemini 2.5 Pro reconstructed the correct financial logic chain, whereas other models defaulted to predicting the masked span without grounding answers in the provided numerical data.

\section{Conclusion and Future Work}

In this paper, we introduced a new dataset and a dynamic framework for evaluating hallucinations of LLMs in the financial domain, providing a nuanced view of their current capabilities. Our findings indicate that even state-of-the-art models struggle with the nuances of financial tabular data. While leading models are approaching the accuracy required for less critical applications, the fundamental challenges of ensuring factual integrity, particularly regarding numerical scale in complex reasoning, remain the primary barrier to their safe deployment in accuracy-critical financial workflows.

For future work, we plan to expand the benchmarking datasets to include more document types and more complex reasoning scenarios. Another promising direction is to study how factors like table size and the number of tables in the context affect hallucination rates.
In summary, we believe that robust, domain-specific evaluation is a crucial step towards building more reliable and trustworthy LLMs for financial applications. 

\section{Limitation}
FAITH provides a systematic framework for assessing intrinsic tabular hallucinations in financial contexts, yet several limitations remain.

Our benchmark draws on public U.S. company filings and curated financial tables. This design ensures data reliability and comparability but limits coverage of diverse reporting styles, regional accounting standards, and sector-specific terminologies. Nevertheless, the framework is readily expandable to other companies, industries, and jurisdictions, provided that structured financial data are available. Future work can extend FAITH to international filings and more unstructured financial disclosures.

When constructing masked instances, we retain only those that three frontier LLMs unanimously identify as answerable. Although a pilot study validates the reliability of this heuristic, it may systematically exclude the most challenging or ambiguous spans—precisely those that elicit disagreement or uncertainty across models. As a result, FAITH may slightly underestimate the intrinsic difficulty of hallucination detection. A systematic examination of these ambiguous, inter-model disagreement cases constitutes a valuable direction for future research, potentially enhancing both dataset representativeness and evaluation robustness.

\begin{acks}
    We gratefully acknowledge the support of National University of Singapore IT for providing access to its high-performance computing resources (Grant: NUSREC-HPC-00001).
\end{acks}

\bibliographystyle{ACM-Reference-Format}
\bibliography{main}


\begin{thebibliography}{30}


\ifx \showCODEN    \undefined \def \showCODEN     #1{\unskip}     \fi
\ifx \showDOI      \undefined \def \showDOI       #1{#1}\fi
\ifx \showISBNx    \undefined \def \showISBNx     #1{\unskip}     \fi
\ifx \showISBNxiii \undefined \def \showISBNxiii  #1{\unskip}     \fi
\ifx \showISSN     \undefined \def \showISSN      #1{\unskip}     \fi
\ifx \showLCCN     \undefined \def \showLCCN      #1{\unskip}     \fi
\ifx \shownote     \undefined \def \shownote      #1{#1}          \fi
\ifx \showarticletitle \undefined \def \showarticletitle #1{#1}   \fi
\ifx \showURL      \undefined \def \showURL       {\relax}        \fi
\providecommand\bibfield[2]{#2}
\providecommand\bibinfo[2]{#2}
\providecommand\natexlab[1]{#1}
\providecommand\showeprint[2][]{arXiv:#2}

\bibitem[Akbar et~al\mbox{.}(2024)]%
        {Akbar2024HalluMeasure}
\bibfield{author}{\bibinfo{person}{Shayan~Ali Akbar}, \bibinfo{person}{Md~Mosharaf Hossain}, \bibinfo{person}{Tess Wood}, \bibinfo{person}{Si{-}Chi Chin}, \bibinfo{person}{Erica Salinas}, \bibinfo{person}{Victor Alvarez}, {and} \bibinfo{person}{Erwin Cornejo}.} \bibinfo{year}{2024}\natexlab{}.
\newblock \showarticletitle{HalluMeasure: Fine-grained Hallucination Measurement Using Chain-of-Thought Reasoning}. In \bibinfo{booktitle}{\emph{Proceedings of the 2024 Conference on Empirical Methods in Natural Language Processing, {EMNLP} 2024}}. \bibinfo{publisher}{Association for Computational Linguistics}, \bibinfo{pages}{15020--15037}.
\newblock
\urldef\tempurl%
\url{https://doi.org/10.18653/V1/2024.EMNLP-MAIN.837}
\showDOI{\tempurl}


\bibitem[Azaria and Mitchell(2023)]%
        {azaria2023internalstatellmknows}
\bibfield{author}{\bibinfo{person}{Amos Azaria} {and} \bibinfo{person}{Tom Mitchell}.} \bibinfo{year}{2023}\natexlab{}.
\newblock \showarticletitle{The Internal State of an {LLM} Knows When It{'}s Lying}. In \bibinfo{booktitle}{\emph{Findings of the Association for Computational Linguistics: EMNLP 2023}}. \bibinfo{publisher}{Association for Computational Linguistics}, \bibinfo{pages}{967--976}.
\newblock
\urldef\tempurl%
\url{https://doi.org/10.18653/v1/2023.findings-emnlp.68}
\showDOI{\tempurl}


\bibitem[Bang et~al\mbox{.}(2025)]%
        {halluLens}
\bibfield{author}{\bibinfo{person}{Yejin Bang}, \bibinfo{person}{Ziwei Ji}, \bibinfo{person}{Alan Schelten}, \bibinfo{person}{Anthony Hartshorn}, \bibinfo{person}{Tara Fowler}, \bibinfo{person}{Cheng Zhang}, \bibinfo{person}{Nicola Cancedda}, {and} \bibinfo{person}{Pascale Fung}.} \bibinfo{year}{2025}\natexlab{}.
\newblock \showarticletitle{{H}allu{L}ens: {LLM} Hallucination Benchmark}. In \bibinfo{booktitle}{\emph{Proceedings of the 63rd Annual Meeting of the Association for Computational Linguistics (Volume 1: Long Papers)}}. \bibinfo{publisher}{Association for Computational Linguistics}, \bibinfo{pages}{24128--24156}.
\newblock
\urldef\tempurl%
\url{https://doi.org/10.48550/ARXIV.2504.17550}
\showDOI{\tempurl}


\bibitem[Cao et~al\mbox{.}(2024)]%
        {cao2024ecc}
\bibfield{author}{\bibinfo{person}{Yupeng Cao}, \bibinfo{person}{Zhi Chen}, \bibinfo{person}{Qingyun Pei}, \bibinfo{person}{Nathan Lee}, \bibinfo{person}{K.~P. Subbalakshmi}, {and} \bibinfo{person}{Papa~Momar Ndiaye}.} \bibinfo{year}{2024}\natexlab{}.
\newblock \showarticletitle{{ECC} Analyzer: Extracting Trading Signal from Earnings Conference Calls using Large Language Model for Stock Volatility Prediction}. In \bibinfo{booktitle}{\emph{Proceedings of the 5th {ACM} International Conference on {AI} in Finance, {ICAIF} 2024}}. \bibinfo{publisher}{{ACM}}, \bibinfo{pages}{257--265}.
\newblock
\urldef\tempurl%
\url{https://doi.org/10.1145/3677052.3698689}
\showDOI{\tempurl}


\bibitem[Chen et~al\mbox{.}(2021)]%
        {Chen2021FINQA}
\bibfield{author}{\bibinfo{person}{Zhiyu Chen}, \bibinfo{person}{Wenhu Chen}, \bibinfo{person}{Charese Smiley}, \bibinfo{person}{Sameena Shah}, \bibinfo{person}{Iana Borova}, \bibinfo{person}{Dylan Langdon}, \bibinfo{person}{Reema Moussa}, \bibinfo{person}{Matt Beane}, \bibinfo{person}{Ting-Hao Huang}, \bibinfo{person}{Bryan Routledge}, {and} \bibinfo{person}{William~Yang Wang}.} \bibinfo{year}{2021}\natexlab{}.
\newblock \showarticletitle{{F}in{QA}: A Dataset of Numerical Reasoning over Financial Data}. In \bibinfo{booktitle}{\emph{Proceedings of the 2021 Conference on Empirical Methods in Natural Language Processing}}. \bibinfo{publisher}{Association for Computational Linguistics}, \bibinfo{pages}{3697--3711}.
\newblock
\urldef\tempurl%
\url{https://doi.org/10.18653/v1/2021.emnlp-main.300}
\showDOI{\tempurl}


\bibitem[Gu et~al\mbox{.}(2024)]%
        {anah2}
\bibfield{author}{\bibinfo{person}{Yuzhe Gu}, \bibinfo{person}{Ziwei Ji}, \bibinfo{person}{Wenwei Zhang}, \bibinfo{person}{Chengqi Lyu}, \bibinfo{person}{Dahua Lin}, {and} \bibinfo{person}{Kai Chen}.} \bibinfo{year}{2024}\natexlab{}.
\newblock \showarticletitle{ANAH-v2: Scaling Analytical Hallucination Annotation of Large Language Models}.
\newblock \bibinfo{journal}{\emph{CoRR}} (\bibinfo{year}{2024}).
\newblock
\urldef\tempurl%
\url{https://doi.org/10.48550/ARXIV.2407.04693}
\showDOI{\tempurl}


\bibitem[Hu et~al\mbox{.}(2024)]%
        {Hu2024RefChecker}
\bibfield{author}{\bibinfo{person}{Xiangkun Hu}, \bibinfo{person}{Dongyu Ru}, \bibinfo{person}{Lin Qiu}, \bibinfo{person}{Qipeng Guo}, \bibinfo{person}{Tianhang Zhang}, \bibinfo{person}{Yang Xu}, \bibinfo{person}{Yun Luo}, \bibinfo{person}{Pengfei Liu}, \bibinfo{person}{Yue Zhang}, {and} \bibinfo{person}{Zheng Zhang}.} \bibinfo{year}{2024}\natexlab{}.
\newblock \showarticletitle{Knowledge-Centric Hallucination Detection}. In \bibinfo{booktitle}{\emph{Proceedings of the 2024 Conference on Empirical Methods in Natural Language Processing}}. \bibinfo{publisher}{Association for Computational Linguistics}, \bibinfo{pages}{6953--6975}.
\newblock
\urldef\tempurl%
\url{https://doi.org/10.18653/v1/2024.emnlp-main.395}
\showDOI{\tempurl}


\bibitem[Kang and Liu(2023)]%
        {deficiencyInFinance}
\bibfield{author}{\bibinfo{person}{Haoqiang Kang} {and} \bibinfo{person}{Xiao{-}Yang Liu}.} \bibinfo{year}{2023}\natexlab{}.
\newblock \showarticletitle{Deficiency of Large Language Models in Finance: An Empirical Examination of Hallucination}.
\newblock \bibinfo{journal}{\emph{CoRR}} (\bibinfo{year}{2023}).
\newblock
\urldef\tempurl%
\url{https://doi.org/10.48550/ARXIV.2311.15548}
\showDOI{\tempurl}


\bibitem[Krumdick et~al\mbox{.}(2024)]%
        {Krumdick2024Bizbench}
\bibfield{author}{\bibinfo{person}{Michael Krumdick}, \bibinfo{person}{Rik Koncel-Kedziorski}, \bibinfo{person}{Viet~Dac Lai}, \bibinfo{person}{Varshini Reddy}, \bibinfo{person}{Charles Lovering}, {and} \bibinfo{person}{Chris Tanner}.} \bibinfo{year}{2024}\natexlab{}.
\newblock \showarticletitle{{B}iz{B}ench: A Quantitative Reasoning Benchmark for Business and Finance}. In \bibinfo{booktitle}{\emph{Proceedings of the 62nd Annual Meeting of the Association for Computational Linguistics (Volume 1: Long Papers)}}. \bibinfo{publisher}{Association for Computational Linguistics}, \bibinfo{pages}{8309--8332}.
\newblock
\urldef\tempurl%
\url{https://doi.org/10.18653/v1/2024.acl-long.452}
\showDOI{\tempurl}


\bibitem[Li et~al\mbox{.}(2023)]%
        {li-etal-2023-halueval}
\bibfield{author}{\bibinfo{person}{Junyi Li}, \bibinfo{person}{Xiaoxue Cheng}, \bibinfo{person}{Xin Zhao}, \bibinfo{person}{Jian-Yun Nie}, {and} \bibinfo{person}{Ji-Rong Wen}.} \bibinfo{year}{2023}\natexlab{}.
\newblock \showarticletitle{{H}alu{E}val: A Large-Scale Hallucination Evaluation Benchmark for Large Language Models}. In \bibinfo{booktitle}{\emph{Proceedings of the 2023 Conference on Empirical Methods in Natural Language Processing}}. \bibinfo{publisher}{Association for Computational Linguistics}, \bibinfo{pages}{6449--6464}.
\newblock
\urldef\tempurl%
\url{https://doi.org/10.18653/v1/2023.emnlp-main.397}
\showDOI{\tempurl}


\bibitem[Lin et~al\mbox{.}(2022)]%
        {lin-etal-2022-truthfulqa}
\bibfield{author}{\bibinfo{person}{Stephanie Lin}, \bibinfo{person}{Jacob Hilton}, {and} \bibinfo{person}{Owain Evans}.} \bibinfo{year}{2022}\natexlab{}.
\newblock \showarticletitle{{T}ruthful{QA}: Measuring How Models Mimic Human Falsehoods}. In \bibinfo{booktitle}{\emph{Proceedings of the 60th Annual Meeting of the Association for Computational Linguistics (Volume 1: Long Papers)}}. \bibinfo{publisher}{Association for Computational Linguistics}, \bibinfo{address}{Dublin, Ireland}, \bibinfo{pages}{3214--3252}.
\newblock
\urldef\tempurl%
\url{https://doi.org/10.18653/v1/2022.acl-long.229}
\showDOI{\tempurl}


\bibitem[Luo et~al\mbox{.}(2024)]%
        {Luo2024HalluDial}
\bibfield{author}{\bibinfo{person}{Wen Luo}, \bibinfo{person}{Tianshu Shen}, \bibinfo{person}{Wei Li}, \bibinfo{person}{Guangyue Peng}, \bibinfo{person}{Richeng Xuan}, \bibinfo{person}{Houfeng Wang}, {and} \bibinfo{person}{Xi Yang}.} \bibinfo{year}{2024}\natexlab{}.
\newblock \showarticletitle{HalluDial: {A} Large-Scale Benchmark for Automatic Dialogue-Level Hallucination Evaluation}.
\newblock \bibinfo{journal}{\emph{CoRR}} (\bibinfo{year}{2024}).
\newblock
\urldef\tempurl%
\url{https://doi.org/10.48550/ARXIV.2406.07070}
\showDOI{\tempurl}


\bibitem[Min et~al\mbox{.}(2023)]%
        {Min2023FActScore}
\bibfield{author}{\bibinfo{person}{Sewon Min}, \bibinfo{person}{Kalpesh Krishna}, \bibinfo{person}{Xinxi Lyu}, \bibinfo{person}{Mike Lewis}, \bibinfo{person}{Wen-tau Yih}, \bibinfo{person}{Pang Koh}, \bibinfo{person}{Mohit Iyyer}, \bibinfo{person}{Luke Zettlemoyer}, {and} \bibinfo{person}{Hannaneh Hajishirzi}.} \bibinfo{year}{2023}\natexlab{}.
\newblock \showarticletitle{{FA}ct{S}core: Fine-grained Atomic Evaluation of Factual Precision in Long Form Text Generation}. In \bibinfo{booktitle}{\emph{Proceedings of the 2023 Conference on Empirical Methods in Natural Language Processing}}. \bibinfo{publisher}{Association for Computational Linguistics}, \bibinfo{pages}{12076--12100}.
\newblock
\urldef\tempurl%
\url{https://doi.org/10.18653/v1/2023.emnlp-main.741}
\showDOI{\tempurl}


\bibitem[Ming et~al\mbox{.}(2024)]%
        {FaithEval}
\bibfield{author}{\bibinfo{person}{Yifei Ming}, \bibinfo{person}{Senthil Purushwalkam}, \bibinfo{person}{Shrey Pandit}, \bibinfo{person}{Zixuan Ke}, \bibinfo{person}{Xuan{-}Phi Nguyen}, \bibinfo{person}{Caiming Xiong}, {and} \bibinfo{person}{Shafiq Joty}.} \bibinfo{year}{2024}\natexlab{}.
\newblock \showarticletitle{FaithEval: Can Your Language Model Stay Faithful to Context, Even If "The Moon is Made of Marshmallows"}.
\newblock \bibinfo{journal}{\emph{CoRR}} (\bibinfo{year}{2024}).
\newblock
\urldef\tempurl%
\url{https://doi.org/10.48550/ARXIV.2410.03727}
\showDOI{\tempurl}


\bibitem[{Monetary Authority of Singapore}(2024)]%
        {MAS2024AI}
\bibfield{author}{\bibinfo{person}{{Monetary Authority of Singapore}}.} \bibinfo{year}{2024}\natexlab{}.
\newblock \showarticletitle{Artificial Intelligence (AI) Model Risk Management}.
\newblock  (\bibinfo{date}{dec} \bibinfo{year}{2024}).
\newblock
\newblock
\shownote{Available at: \url{https://www.mas.gov.sg/-/media/mas-media-library/publications/monographs-or-information-paper/imd/2024/information-paper-on-ai-risk-management-final.pdf}}.


\bibitem[Murtaza et~al\mbox{.}(2025)]%
        {murtaza2025implementing}
\bibfield{author}{\bibinfo{person}{Syed~Shariyar Murtaza}, \bibinfo{person}{Yifan Nie}, \bibinfo{person}{Elias Avan}, \bibinfo{person}{Utkarsh Soni}, \bibinfo{person}{Wanyu Liao}, \bibinfo{person}{Adam Carnegie}, \bibinfo{person}{Cyril~John Mathias}, \bibinfo{person}{Junlin Jiang}, {and} \bibinfo{person}{Eugene Wen}.} \bibinfo{year}{2025}\natexlab{}.
\newblock \showarticletitle{Implementing Retrieval Augmented Generation Technique on Unstructured and Structured Data Sources in a Call Center of a Large Financial Institution}. In \bibinfo{booktitle}{\emph{Proceedings of the 2025 Conference of the Nations of the Americas Chapter of the Association for Computational Linguistics: Human Language Technologies, {NAACL} 2025 - Volume 3: Industry Track, Albuquerque, New Mexico, USA, April 30, 2025}}. \bibinfo{publisher}{Association for Computational Linguistics}, \bibinfo{pages}{598--606}.
\newblock
\urldef\tempurl%
\url{https://doi.org/10.18653/V1/2025.NAACL-INDUSTRY.48}
\showDOI{\tempurl}


\bibitem[Niu et~al\mbox{.}(2024)]%
        {RAGTruth}
\bibfield{author}{\bibinfo{person}{Cheng Niu}, \bibinfo{person}{Yuanhao Wu}, \bibinfo{person}{Juno Zhu}, \bibinfo{person}{Siliang Xu}, \bibinfo{person}{KaShun Shum}, \bibinfo{person}{Randy Zhong}, \bibinfo{person}{Juntong Song}, {and} \bibinfo{person}{Tong Zhang}.} \bibinfo{year}{2024}\natexlab{}.
\newblock \showarticletitle{{RAGT}ruth: A Hallucination Corpus for Developing Trustworthy Retrieval-Augmented Language Models}. In \bibinfo{booktitle}{\emph{Proceedings of the 62nd Annual Meeting of the Association for Computational Linguistics (Volume 1: Long Papers)}}. \bibinfo{publisher}{Association for Computational Linguistics}, \bibinfo{pages}{10862--10878}.
\newblock
\urldef\tempurl%
\url{https://doi.org/10.18653/v1/2024.acl-long.585}
\showDOI{\tempurl}


\bibitem[Oh et~al\mbox{.}(2024)]%
        {Oh2024ERBench}
\bibfield{author}{\bibinfo{person}{Jio Oh}, \bibinfo{person}{Soyeon Kim}, \bibinfo{person}{Junseok Seo}, \bibinfo{person}{Jindong Wang}, \bibinfo{person}{Ruochen Xu}, \bibinfo{person}{Xing Xie}, {and} \bibinfo{person}{Steven~Euijong Whang}.} \bibinfo{year}{2024}\natexlab{}.
\newblock \showarticletitle{ERBench: An Entity-Relationship based Automatically Verifiable Hallucination Benchmark for Large Language Models}.
\newblock \bibinfo{journal}{\emph{CoRR}} (\bibinfo{year}{2024}).
\newblock
\urldef\tempurl%
\url{https://doi.org/10.48550/ARXIV.2403.05266}
\showDOI{\tempurl}


\bibitem[Roychowdhury(2024)]%
        {halluMiniJourney}
\bibfield{author}{\bibinfo{person}{Sohini Roychowdhury}.} \bibinfo{year}{2024}\natexlab{}.
\newblock \showarticletitle{Journey of Hallucination-minimized Generative AI Solutions for Financial Decision Makers}. In \bibinfo{booktitle}{\emph{Proceedings of the 17th ACM International Conference on Web Search and Data Mining}} \emph{(\bibinfo{series}{WSDM '24})}. \bibinfo{publisher}{Association for Computing Machinery}, \bibinfo{pages}{1180–1181}.
\newblock
\urldef\tempurl%
\url{https://doi.org/10.1145/3616855.3635737}
\showDOI{\tempurl}


\bibitem[Roychowdhury et~al\mbox{.}(2023)]%
        {HalluMiniFram}
\bibfield{author}{\bibinfo{person}{Sohini Roychowdhury}, \bibinfo{person}{Andres Alvarez}, \bibinfo{person}{Brian Moore}, \bibinfo{person}{Marko Krema}, \bibinfo{person}{Maria~Paz Gelpi}, \bibinfo{person}{Punit Agrawal}, \bibinfo{person}{Federico~Martin Rodriguez}, \bibinfo{person}{Angel Rodriguez}, \bibinfo{person}{Jose~Ramon Cabrejas}, \bibinfo{person}{Pablo~Martinez Serrano}, {and} \bibinfo{person}{Arijit Mukherjee}.} \bibinfo{year}{2023}\natexlab{}.
\newblock \showarticletitle{{ Hallucination-minimized Data-to-answer Framework for Financial Decision-makers }}. In \bibinfo{booktitle}{\emph{2023 IEEE International Conference on Big Data (BigData)}}. \bibinfo{publisher}{IEEE Computer Society}, \bibinfo{pages}{4693--4702}.
\newblock
\urldef\tempurl%
\url{https://doi.org/10.1109/BigData59044.2023.10386232}
\showDOI{\tempurl}


\bibitem[Sarmah et~al\mbox{.}(2024)]%
        {reduHalluFin}
\bibfield{author}{\bibinfo{person}{Bhaskarjit Sarmah}, \bibinfo{person}{Dhagash Mehta}, \bibinfo{person}{Stefano Pasquali}, {and} \bibinfo{person}{Tianjie Zhu}.} \bibinfo{year}{2024}\natexlab{}.
\newblock \showarticletitle{Towards reducing hallucination in extracting information from financial reports using Large Language Models}. In \bibinfo{booktitle}{\emph{Proceedings of the Third International Conference on AI-ML Systems}} \emph{(\bibinfo{series}{AIMLSystems '23})}. \bibinfo{publisher}{Association for Computing Machinery}, Article \bibinfo{articleno}{39}.
\newblock
\urldef\tempurl%
\url{https://doi.org/10.1145/3639856.3639895}
\showDOI{\tempurl}


\bibitem[Shah et~al\mbox{.}(2025)]%
        {beyondCutoff}
\bibfield{author}{\bibinfo{person}{Agam Shah}, \bibinfo{person}{Liqin Ye}, \bibinfo{person}{Sebastian Jaskowski}, \bibinfo{person}{Wei Xu}, {and} \bibinfo{person}{Sudheer Chava}.} \bibinfo{year}{2025}\natexlab{}.
\newblock \showarticletitle{Beyond the Reported Cutoff: Where Large Language Models Fall Short on Financial Knowledge}.
\newblock \bibinfo{journal}{\emph{CoRR}} (\bibinfo{year}{2025}).
\newblock
\urldef\tempurl%
\url{https://doi.org/10.48550/ARXIV.2504.00042}
\showDOI{\tempurl}


\bibitem[Wang et~al\mbox{.}(2025)]%
        {Wang2025FinTagging}
\bibfield{author}{\bibinfo{person}{Yan Wang}, \bibinfo{person}{Yang Ren}, \bibinfo{person}{Lingfei Qian}, \bibinfo{person}{Xueqing Peng}, \bibinfo{person}{Keyi Wang}, \bibinfo{person}{Yi Han}, \bibinfo{person}{Dongji Feng}, \bibinfo{person}{Xiao{-}Yang Liu}, \bibinfo{person}{Jimin Huang}, {and} \bibinfo{person}{Qianqian Xie}.} \bibinfo{year}{2025}\natexlab{}.
\newblock \showarticletitle{FinTagging: An LLM-ready Benchmark for Extracting and Structuring Financial Information}.
\newblock \bibinfo{journal}{\emph{CoRR}} (\bibinfo{year}{2025}).
\newblock
\urldef\tempurl%
\url{https://doi.org/10.48550/ARXIV.2505.20650}
\showDOI{\tempurl}


\bibitem[Wu et~al\mbox{.}(2025b)]%
        {Wu2025RealHiTBench}
\bibfield{author}{\bibinfo{person}{Pengzuo Wu}, \bibinfo{person}{Yuhang Yang}, \bibinfo{person}{Guangcheng Zhu}, \bibinfo{person}{Chao Ye}, \bibinfo{person}{Hong Gu}, \bibinfo{person}{Xu Lu}, \bibinfo{person}{Ruixuan Xiao}, \bibinfo{person}{Bowen Bao}, \bibinfo{person}{Yijing He}, \bibinfo{person}{Liangyu Zha}, \bibinfo{person}{Wentao Ye}, \bibinfo{person}{Junbo Zhao}, {and} \bibinfo{person}{Haobo Wang}.} \bibinfo{year}{2025}\natexlab{b}.
\newblock \showarticletitle{RealHiTBench: {A} Comprehensive Realistic Hierarchical Table Benchmark for Evaluating LLM-Based Table Analysis}.
\newblock \bibinfo{journal}{\emph{CoRR}} (\bibinfo{year}{2025}).
\newblock
\urldef\tempurl%
\url{https://doi.org/10.48550/ARXIV.2506.13405}
\showDOI{\tempurl}


\bibitem[Wu et~al\mbox{.}(2025a)]%
        {Wu2025TableBench}
\bibfield{author}{\bibinfo{person}{Xianjie Wu}, \bibinfo{person}{Jian Yang}, \bibinfo{person}{Linzheng Chai}, \bibinfo{person}{Ge Zhang}, \bibinfo{person}{Jiaheng Liu}, \bibinfo{person}{Xeron Du}, \bibinfo{person}{Di Liang}, \bibinfo{person}{Daixin Shu}, \bibinfo{person}{Xianfu Cheng}, \bibinfo{person}{Tianzhen Sun}, \bibinfo{person}{Tongliang Li}, \bibinfo{person}{Zhoujun Li}, {and} \bibinfo{person}{Guanglin Niu}.} \bibinfo{year}{2025}\natexlab{a}.
\newblock \showarticletitle{TableBench: {A} Comprehensive and Complex Benchmark for Table Question Answering}. In \bibinfo{booktitle}{\emph{AAAI-25, Sponsored by the Association for the Advancement of Artificial Intelligence, February 25 - March 4, 2025, Philadelphia, PA, {USA}}}. \bibinfo{publisher}{{AAAI} Press}, \bibinfo{pages}{25497--25506}.
\newblock
\urldef\tempurl%
\url{https://doi.org/10.1609/AAAI.V39I24.34739}
\showDOI{\tempurl}


\bibitem[Yu et~al\mbox{.}(2024)]%
        {Yu2024ReEval}
\bibfield{author}{\bibinfo{person}{Xiaodong Yu}, \bibinfo{person}{Hao Cheng}, \bibinfo{person}{Xiaodong Liu}, \bibinfo{person}{Dan Roth}, {and} \bibinfo{person}{Jianfeng Gao}.} \bibinfo{year}{2024}\natexlab{}.
\newblock \showarticletitle{ReEval: Automatic Hallucination Evaluation for Retrieval-Augmented Large Language Models via Transferable Adversarial Attacks}. In \bibinfo{booktitle}{\emph{Findings of the Association for Computational Linguistics: {NAACL} 2024, Mexico City, Mexico, June 16-21, 2024}}. \bibinfo{publisher}{Association for Computational Linguistics}, \bibinfo{pages}{1333--1351}.
\newblock
\urldef\tempurl%
\url{https://doi.org/10.18653/V1/2024.FINDINGS-NAACL.85}
\showDOI{\tempurl}


\bibitem[Zhao et~al\mbox{.}(2022)]%
        {zhao2022multihiertt}
\bibfield{author}{\bibinfo{person}{Yilun Zhao}, \bibinfo{person}{Yunxiang Li}, \bibinfo{person}{Chenying Li}, {and} \bibinfo{person}{Rui Zhang}.} \bibinfo{year}{2022}\natexlab{}.
\newblock \showarticletitle{{M}ulti{H}iertt: Numerical Reasoning over Multi Hierarchical Tabular and Textual Data}. In \bibinfo{booktitle}{\emph{Proceedings of the 60th Annual Meeting of the Association for Computational Linguistics (Volume 1: Long Papers)}}. \bibinfo{publisher}{Association for Computational Linguistics}, \bibinfo{pages}{6588--6600}.
\newblock
\urldef\tempurl%
\url{https://doi.org/10.18653/v1/2022.acl-long.454}
\showDOI{\tempurl}


\bibitem[Zheng et~al\mbox{.}(2023)]%
        {Zheng2023Judging}
\bibfield{author}{\bibinfo{person}{Lianmin Zheng}, \bibinfo{person}{Wei{-}Lin Chiang}, \bibinfo{person}{Ying Sheng}, \bibinfo{person}{Siyuan Zhuang}, \bibinfo{person}{Zhanghao Wu}, \bibinfo{person}{Yonghao Zhuang}, \bibinfo{person}{Zi Lin}, \bibinfo{person}{Zhuohan Li}, \bibinfo{person}{Dacheng Li}, \bibinfo{person}{Eric~P. Xing}, \bibinfo{person}{Hao Zhang}, \bibinfo{person}{Joseph~E. Gonzalez}, {and} \bibinfo{person}{Ion Stoica}.} \bibinfo{year}{2023}\natexlab{}.
\newblock \showarticletitle{Judging LLM-as-a-judge with MT-Bench and Chatbot Arena}.
\newblock \bibinfo{journal}{\emph{CoRR}} (\bibinfo{year}{2023}).
\newblock
\urldef\tempurl%
\url{https://doi.org/10.48550/ARXIV.2306.05685}
\showDOI{\tempurl}


\bibitem[Zheng et~al\mbox{.}(2024)]%
        {llamafactory}
\bibfield{author}{\bibinfo{person}{Yaowei Zheng}, \bibinfo{person}{Richong Zhang}, \bibinfo{person}{Junhao Zhang}, \bibinfo{person}{Yanhan Ye}, {and} \bibinfo{person}{Zheyan Luo}.} \bibinfo{year}{2024}\natexlab{}.
\newblock \showarticletitle{{L}lama{F}actory: Unified Efficient Fine-Tuning of 100+ Language Models}. In \bibinfo{booktitle}{\emph{Proceedings of the 62nd Annual Meeting of the Association for Computational Linguistics (Volume 3: System Demonstrations)}}. \bibinfo{publisher}{Association for Computational Linguistics}, \bibinfo{pages}{400--410}.
\newblock
\urldef\tempurl%
\url{https://doi.org/10.18653/v1/2024.acl-demos.38}
\showDOI{\tempurl}


\bibitem[Zhu et~al\mbox{.}(2021)]%
        {Zhu2021TATQA}
\bibfield{author}{\bibinfo{person}{Fengbin Zhu}, \bibinfo{person}{Wenqiang Lei}, \bibinfo{person}{Youcheng Huang}, \bibinfo{person}{Chao Wang}, \bibinfo{person}{Shuo Zhang}, \bibinfo{person}{Jiancheng Lv}, \bibinfo{person}{Fuli Feng}, {and} \bibinfo{person}{Tat-Seng Chua}.} \bibinfo{year}{2021}\natexlab{}.
\newblock \showarticletitle{{TAT}-{QA}: A Question Answering Benchmark on a Hybrid of Tabular and Textual Content in Finance}. In \bibinfo{booktitle}{\emph{Proceedings of the 59th Annual Meeting of the Association for Computational Linguistics and the 11th International Joint Conference on Natural Language Processing (Volume 1: Long Papers)}}. \bibinfo{publisher}{Association for Computational Linguistics}, \bibinfo{pages}{3277--3287}.
\newblock
\urldef\tempurl%
\url{https://doi.org/10.18653/v1/2021.acl-long.254}
\showDOI{\tempurl}


\end{thebibliography}

\appendix

\section{Prompt Templates}

\subsection{Financial Span Answerability Annotation}
\label{sec:ans_prompt}

\begingroup
\sloppy
\begin{small}

\lstset{basicstyle=\ttfamily\small,columns=fullflexible,breaklines=true,frame=single,showstringspaces=false}
\begin{spacing}{0.68}
You are given the following tables from a 10-K annual report: \\{\texttt{\{tables\_with\_pretext\}}} \\
Filing date: {\texttt{\{filing\_date\}}} \\
Sentence: {\texttt{\{pre\_sentence\}}} {\texttt{\{sentence\}}} {\texttt{\{post\_sentence\}}} \\\\
Your task: \\
For each highlighted span substring (shown as \texttt{{[span]}}) in the sentence, determine if the span is:
\begin{itemize}[label=-]
    \item \textbf{unanswerable}: Spans that
    \begin{itemize}[label=-]
        \item do not represent some type of numeric financial data (e.g., phone numbers, pincodes, or any other noise), or
        \item the span cannot be derived from or supported by the table by any method.
    \end{itemize}
    \item \textbf{answerable}: Spans that
        \begin{itemize}[label=-]
        \item can be directly found in the table, or
        \item can be derived through some calculations (such as addition, subtraction, multiplication, or division), or
        \item can be inferred via deeper reasoning involving multiple table entries
    \end{itemize}
\end{itemize}
\textbf{Instructions:}
\begin{enumerate}[nosep]
    \item Carefully analyze every given table and the given sentence.
    \item For each \texttt{{span}} in the sentence, label the {{span}} as \textit{answerable} or \textit{unanswerable}.
    \item Provide a concise explanation for your reasoning.
    \end{enumerate}
\textbf{Output Format:}
\begin{verbatim}
```json
{
    "reasoning": "<A detailed explanation for each highlighted
        span in the sentence — if it is answerable: how it can
        be matched, derived, or inferred from the table; or if
        it is unanswerable: a brief reason why.>",
    "spans": {
        "<span1>": "answerable" | "unanswerable",
        ...
    }
}
```\end{verbatim}
\end{spacing}
\end{small}
\endgroup

\vspace{1em}

\label{sec:pred_prompt}
\subsection{Financial Metric Prediction}

\begingroup
\sloppy
\begin{small}

\lstset{basicstyle=\ttfamily\small,columns=fullflexible,breaklines=true,frame=single,showstringspaces=false}

\begin{spacing}{0.7}
You are a financial analyst. \\
Your task is to fill in the masked blank in a sentence using the provided financial data tables. \\
Hint: The masked content is a single positive value that fits the context \texttt{\{unit\_description\}}.\\\\
\textbf{Instructions:}
\begin{enumerate}[nosep,leftmargin=*]
    \item \textbf{Analyze the Request:} Carefully read the sentence and examine the provided tables to understand what information is needed to fill in the blank.
    \item \textbf{Reason Step-by-Step:} Before providing the final answer, write out your reasoning process. Explain how you will find the value, including any calculations.
    \begin{itemize}[label=-]
        \item If you are extracting a value, mention which table and cell you are getting it from.
        \item If you are performing a calculation, show the formula and the values you are using.
    \end{itemize}
    \item \textbf{Categorize Your Reasoning:} Based on your reasoning, classify it into one of the following scenarios:
    \begin{itemize}[label=-]
        \item \textbf{A. Direct Lookup:} The value is directly extracted from a single cell in a table.
        \item \textbf{B. Simple Calculation (Single Metric):} The result is calculated from a single metric across different time periods, categories, or rows (e.g., calculating a year-over-year change).
        \item \textbf{C. Simple Calculation (Two Metrics):} The result is calculated using two different metrics (e.g., calculating a ratio).
        \item \textbf{D. Complex Calculation:} The reasoning involves more than two metrics or multiple complex steps.
    \end{itemize}
    \item \textbf{Format the Final Answer:} After your reasoning, provide the final answer in a JSON block with the following structure.
\end{enumerate}
\textbf{JSON Output Format:}
\begin{verbatim}
```json
{
  "results": {
    "answer": "<The calculated or extracted value>",
    "scenario": "<A, B, C, or D>",
    "necessary_metrics": ["<metric_name_1>", "..."],
    "reference": ["<table_identifier_1>", "..."]
  }
}
\end{verbatim}
\textbf{Field Explanations:}
\begin{itemize}[label=-,leftmargin=*]
    \item \texttt{answer}: The value to fill in the masked blank. Format it professionally (i.e., with rounding and units, etc.).
    \item \texttt{scenario}: One of A'', B'', C'', or D'' based on your reasoning.
    \item \texttt{necessary\_metrics}: A list of all metric names from the tables required to derive the answer.
    \item \texttt{reference}: A list of all table identifiers for the tables used.
\end{itemize}
\textbf{Inputs:}
Tables: {\texttt{\{tables\_with\_pretext\}}} \\
Sentence: {\texttt{\{pre\_sentence\}}} {\texttt{\{sentence\}}} {\texttt{\{post\_sentence\}}}
\end{spacing}
\end{small}
\endgroup

\end{document}